\documentclass{article} 
\usepackage{iclr2025_conference,times}

\usepackage{amsmath,amsfonts,bm}

\def\eqref#1{equation~\ref{#1}}

\def\1{\bm{1}}

\DeclareMathAlphabet{\mathsfit}{\encodingdefault}{\sfdefault}{m}{sl}
\SetMathAlphabet{\mathsfit}{bold}{\encodingdefault}{\sfdefault}{bx}{n}

\usepackage{hyperref}
\usepackage{url}
\usepackage{graphicx} 
\graphicspath{{figs/}}
\usepackage{natbib}
\usepackage{xcolor}
\usepackage{makecell}
\usepackage{chngcntr}
\usepackage{paralist}
\counterwithout{figure}{section}

\usepackage{booktabs}  

\newcommand{\coolcampfire}{\textbf{Campfire}}

\usepackage{placeins}

 \iclrfinalcopy

\title{Out-of-distribution evaluations of channel agnostic masked autoencoders \\  in fluorescence microscopy }

\author{Christian John Hurry\thanks{Corresponding author, christian.j.hurry@gsk.com  }, Jinjie Zhang,
Olubukola Ishola, Emma Slade, Cuong Q. Nguyen\\
GSK AIML  \\
79 New Oxford Street, London, United Kingdom, WC1A 1DG
}

\begin{document}

\maketitle

\begin{abstract}
Developing computer vision for high-content screening is challenging due to various sources of distribution-shift caused by changes in experimental conditions, perturbagens, and fluorescent markers. 
The impact of different sources of distribution-shift are confounded in typical evaluations of models based on transfer learning, which limits interpretations of how changes to model design and training affect generalisation. 
We propose an evaluation scheme that isolates sources of distribution-shift using the JUMP-CP dataset, allowing researchers to evaluate generalisation with respect to specific sources of distribution-shift.
We then present a channel-agnostic masked autoencoder \coolcampfire{} which, via a shared decoder for all channels, scales effectively to datasets containing many different fluorescent markers, and show that it generalises to out-of-distribution experimental batches, perturbagens, and fluorescent markers, and also demonstrates successful transfer learning from one cell type to another.
\end{abstract}


\section{Introduction}

Phenotypic drug discovery, in which cells or animal models are subject to a perturbation and monitored for a desired change in phenotype, has seen a resurgence due to its success in finding compounds that meet regulatory approval \citep{zheng2013phenotypic,boutros2015microscopy,zanella2010high}. To quantify the effect of perturbations, it is common to use high content screening (HCS), a method in which batches of cells are stimulated with thousands of compounds in parallel, and multiple markers of changes in phenotype are measured simultaneously. In comparison with modalities based on sequencing technologies, imaging is more time- and cost-effective at scale and has been the main modality of HCS data. This necessitated the development of automated pipelines that extract biologically relevant features from cellular imaging data. Typically, this has involved traditional methods based on cell-segmentation and feature extraction and has been applied in various applications including protein sub-cellular localisation \citep{parnamaa2017accurate},  quantitative structure-activity relationship modelling \citep{nguyen2023molecule} and identifying mechanism of action \citep{durr2016single,wong2023deep} and markers of drug resistance \citep{kelley2023high}. 

In recent years, the volume of high content imaging (HCI) data has increased in scale due to the introduction of the Cell Painting assay \citep{bray2016cell, seal2024cell}. This assay provides standardisation for HCI by using a specific set of dyes, optimised to highlight and contrast several cellular compartments of interest \citep{cimini2023optimizing}. The public JUMP-CP dataset was generated using the Cell Painting assay, and contains images of millions of cells, subject to 116,000 compound and 22,000 genetic perturbations, derived from multiple laboratories and institutions \citep{chandrasekaran2023jump}. Examples of the application of the JUMP-CP dataset to-date include predicting the mechanism of action of kinase inhibitors \citep{dee2024cell}, quantifying differences in data derived from different experimental batches \citep{arevalo2024evaluating}, and predicting the molecular structure of perturbagens from cell imaging \citep{watkinson2024weakly}.

Recently, large-scale public datasets have been leveraged to produce SOTA \textit{foundation} models in various biological domains, including single cell sequencing data \citep{yang2022scbert,cui2024scgpt,heimberg2024cell}, protein and complex structure prediction \citep{abramson2024accurate}, and pathology \citep{vorontsov2023virchow,juyal2024pluto,dippel2024rudolfv}. Foundation models differ from typical deep neural networks as they learn representations which are useful across a range of biologically relevant tasks. To achieve this, they are typically trained in self-supervised fashion, such that representations are learned in a manner which is less biased to a particular task in comparison with fully supervised methods. Recent work has applied self-supervised learning to HCI \citep{borowa2024decoding,bourriez2024chada,bao2023contextual,kraus2024masked} and has shown that models scale with dataset and model size suggesting a route towards a foundation model for HCI \citep{kraus2024masked}. 

It remains an open challenge to develop a foundation model for HCI, in part due to the variety of sources of distribution-shift that are common to fluorescence microscopy. This includes changes in experimental conditions, instruments of measurement, and the cell types and perturbations under consideration. Another source of distribution-shift that is a particular challenge with HCI is changes to the set of fluorescent markers used to generate images. These differ between experimental screens to highlight the most relevant cellular compartments. The evaluation of models for HCI with respect to different sources of distribution-shift, varies between previous works. In some cases, models are shown to generalise to different microscopy screens, which can include changes in cell types, fluorescent markers and perturbations in combination and is, therefore, the most challenging form of generalisation. This, however, confounds all sources of distribution-shift, making it difficult to assess how changes in model architecture or training protocol affect generalisation to distribution-shift of a specific type. Therefore, to assess and benchmark different models for HCI, it is important to show \textit{explicit} generalisation to \begin{inparaenum}[\itshape i\upshape)]
\item new experimental plates,
\item changes in perturbagens,
\item changes in fluorescent markers, and
\item new microscopy screens. 
\end{inparaenum}
By isolating different sources of distribution-shift, we can provide a more comprehensive analysis of the universality of any given model for HCI.

To address these generalisation requirements we introduce \coolcampfire{}\footnote{Code and model checkpoint can be found at \urlstyle{same} \url{https://github.com/GSK-AI/campfire}.} (\textbf{C}hannel \textbf{a}gnostic \textbf{m}orphological \textbf{p}rofiling \textbf{f}rom \textbf{i}maging under a \textbf{r}ange of \textbf{e}xperimental settings). This is a channel-agnostic masked autoencoder that uses a shared decoder for all channels, such that it scales more effectively to datasets containing a wide variety of fluorescent markers, and that can handle out-of-distribution fluorescent markers during inference. In this work, we propose an evaluation scheme based on the JUMP-CP dataset, that isolates sources of distribution-shift common to fluorescence microscopy, and demonstrate that \coolcampfire{} learns representations of data that generalise across plates, perturbagens, fluorescent markers, and cell type. 

\subsection{Related work}

To address the challenges of HCI, several works have recently adapted vision transformers (ViT) which can accept input sequences of varying length during training and inference \citep{dosovitskiy2020image}. Each channel in HCI provides information from a distinct cellular compartment. To utilise this ChannelViT \citep{bao2023channel} and ChAda-ViT \citep{bourriez2024chada} proposed encoding each channel of an image into a set of distinct patch embeddings by applying a shared projection layer to each channel.
These patch embeddings were encoded with the channel of origin via a set of learnable embeddings, akin to position encoding. Notably, self-supervised ChAda-ViT demonstrated successful transfer learning from one HCI dataset to another. CA-MAE \citep{kraus2024masked} was also trained in self-supervised fashion using HCI but was trained to reconstruct images of wells from masked input with arbitrary combinations of channels. This work demonstrated that masked autoencoders (MAEs) \citep{he2022masked} scale, such that increasing the amount of available HCI and compute lead to better image representations. While not channel-agnostic, ContextViT \citep{bao2023contextual} showed improved performance on out-of-distribution experimental plates, by appending an additional \textit{context} embedding to the sequence of patch embeddings. Context embeddings were formed by projecting the average patch embedding through a linear layer. To the best of our knowledge, a model has not been explicitly designed to generalise to OOD fluorescent markers.


\section{Methodology}

\subsection{JUMP-CP dataset}

In this work we used Source 3 of the JUMP-CP dataset \cite{chandrasekaran2023jump}, consisting of images of cells perturbed by small molecules. We considered all 25 TARGET2 and 237 COMPOUND plates, comprised of 384 wells each. To avoid data leakage, we assigned images to training, validation or test sets according to the well from which they were derived. This assignment differed between the TARGET2 and COMPOUND plates. 

\begin{figure*}[t]
\vskip 0.2in
\begin{center}
\centerline{\includegraphics[width=0.7\linewidth]{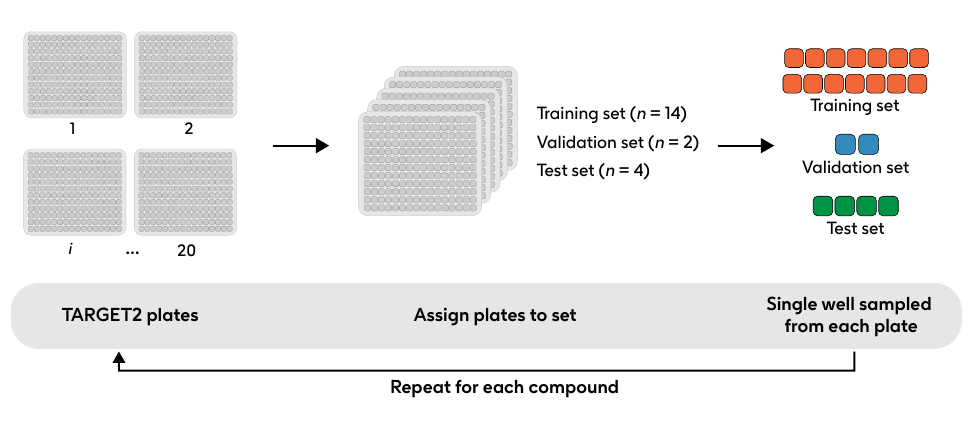}}
  \caption{A sketch of how wells are assigned to training/validation/test sets for a given compound in TARGET2 plates. For each compound, we select 14 plates from which to sample a single well treated with that compound for the training set, 2 plates to sample a well for the validation set, and 4 plates for the test set. The exception are 60 randomly selected compounds which are all held-out of training. We also hold 5 TARGET2 plates out of training, with all wells added to the test set. }
  \label{fig: target split}
\end{center}
\vskip -0.2in
\end{figure*}

Each TARGET2 plate contained wells stimulated by 302 compounds, including 8 positive controls and 1 negative control. We randomly selected 60 compounds to hold out of training, excluding the controls. We also held 5 TARGET2 plates out of training, such that we had 20 in-distribution (ID) and 5 out-of-distribution (OOD) plates. As sketched in Fig. \ref{fig: target split} for each compound we selected 14 plates for training, 2 plates for validation, and 4 plates for testing, sampling a single well from each of these plates for that compound. Consequently, our training/validation/test set was represented by 14/2/4 wells for each compound, where the plates from which these wells were derived may have differed for each compound. Wells within the test set were either 
 \begin{inparaenum}[\itshape i\upshape)]
\item from an ID plate, with ID compound 
\item from an OOD plate, with ID compound 
\item from an ID plate, with OOD compound 
\item from an OOD plate, with OOD compound. 
\end{inparaenum}
Hence, at the cost of less images being included in our training set, this data splitting procedure allowed allowed us to evaluate our model against different sources of distribution-shift, by evaluating performance on different subsets of our test set, isolating the effect of distribution-shift caused by plate of origin or whether or not the compound was seen during training. 

We included additional wells in the training set from COMPOUND plates, which in total represented cells stimulated by 58456 different compounds. For these plates, we assigned each well to the training set with probability $p_{t} = 0.5$ to limit the number of single cell images in our dataset. To produce a consistent evaluation scheme, we did not include COMPOUND plates in our test or validation sets.

\subsection{\coolcampfire{}: a channel agnostic masked autoencoder with channel embeddings and channel subsampling}

\begin{figure*}[t]
\vskip 0.2in
\begin{center}
\centerline{\includegraphics[width=0.65\linewidth]{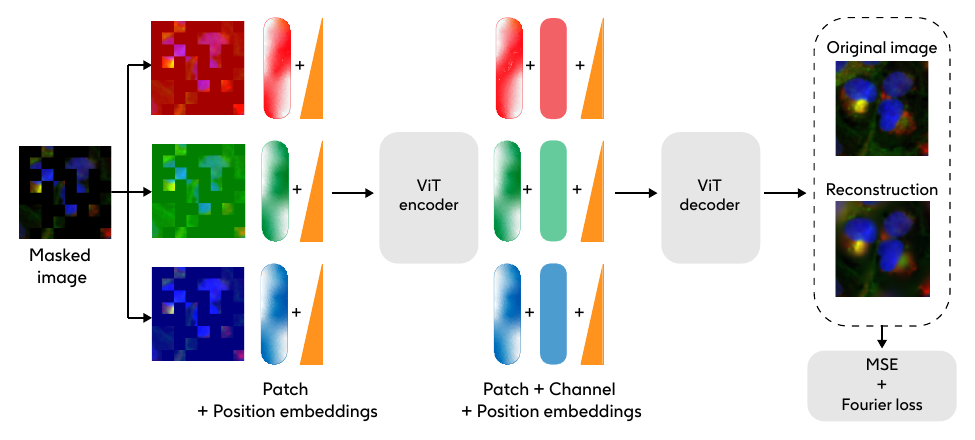}}
  \caption{A sketch of the \coolcampfire{} architecture. }
  \label{fig: campfire sketch}
\end{center}
\vskip -0.2in
\end{figure*}

 \coolcampfire{}, sketched in Fig. \ref{fig: campfire sketch}, is a channel-agnostic MAE. Well images were preprocessed into cell-centred tiles of size (112,112) for model input. These were further processed by a \textit{3D convolutional layer}, with kernel and stride of shape $(1,P,P)$, where $P$ was the size of a patch, to form $N \times C$ embeddings of dimension $D$, each representing $N$ non-overlapping $(P,P)$ patches for each channel of a cell-centred tile with $C$ channels. Applying the same convolution to each channel, mapped image patches to embeddings in a channel agnostic manner, such that the model could be trained with images of inconsistent channel ordering, number, or type. Patch embeddings were augmented by adding sinusoidal and RoPE position embeddings \citep{su2024roformer,heo2025rotary}. 

The sequence of position encoded patch embeddings were then fed through an asymmetric MAE. A random subset of patch positions indexed $1,\dots,N$ were selected to be masked, with $p_m$ denoting the fraction of masked patches. For all channels, patch embeddings corresponding to the masked positions were removed from input to the encoder. The latent state of the sequence of embeddings was then projected through a linear layer to the smaller embedding dimension of the decoder. This sequence was padded with mask tokens, each representing a patch embedding removed during masking. Each patch embedding was augmented with sinusoidal and RoPE position embeddings, and an additional channel embedding, to encode the channel of origin in the patch. To compute the channel embedding, we averaged all patch embeddings in the batch from the same channel, and projected this through a linear layer shared by all channels. This linear layer, therefore, was trained to produce channel embeddings, given the average patch embedding from that channel.  

The latent state of the input to the decoder was passed to an objective function, detailed in App. \ref{app: loss}, to optimise the reconstruction of the original input images from the masked input. Lastly, at inference time, the decoder of our model was discarded, the full unmasked set of patch embeddings were fed-forward through the encoder, and the average of the latent state of the patch embeddings was used as our final representation of a cell-centred tile.

\section{Results}

\subsection{Model training and reconstruction of cell-centred tiles}

\coolcampfire{} was trained on 16 A100 (80GB) GPUs for 1920 GPU hours, with a per-GPU batch size of 400, for 50 epochs. The patch size was set to $(14,14)$  and the fraction of patches that were masked from training was set to $p_m = 0.8$. During training we used random flipping, rotation and channel normalisation as additional augmentations. We used the AdamW optimiser with learning rate $lr=5 \times 10^{-4}$ with weight decay $wd=0.005$, and applied cosine annealing with 20 epochs of linear warm up with a starting learning rate $lr=1 \times 10^{-5}$ and minimum learning rate of $\eta=1\times10^{-6}$. These hyperparameters were chosen as the result of an ablation study, detailed in App. \ref{app: ablation}.

\begin{figure*}[t]
\vskip 0.2in
\begin{center}
\includegraphics[width=0.85\linewidth]{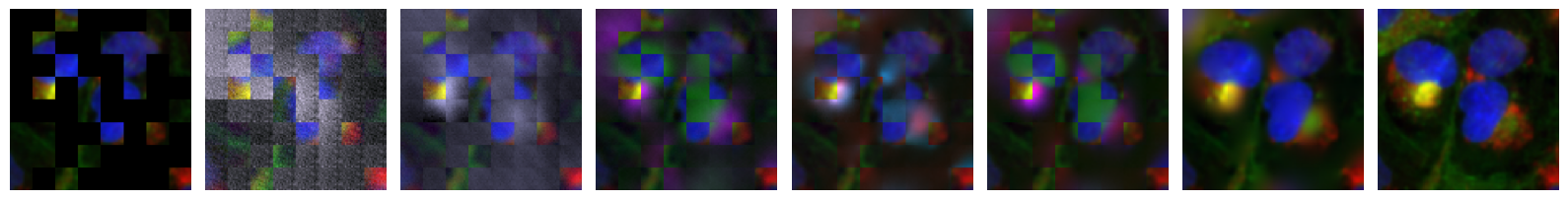}
  \caption{An example of the reconstruction of a cell-centred tile. From left to right, we show the masked input, followed by six reconstructed images at different epochs, with epoch increasing from left to right, and lastly the original cell-centred tile to be reconstructed. }
  \label{fig: visualisation}
\end{center}
\vskip -0.2in
\end{figure*}

The images of the JUMP-CP dataset consisted of 5 fluorescent channels which highlighted the Nucleus (Nu), Actin+Golgi Apparatus+Plasma Membrane (Ac), Mitochondria (M), Endoplasmic Reticulum (ER), and the Nucleolus+cytoplasmic RNA (cyRNA). 
We trained our model with the Nu channel and two additional channels, Ac and M, selected at random. The ER and cyRNA channels were held-out to evaluate OOD performance. For each batch in training, we sampled a subset of channels, $S$, from the set of all available channels, with uniform probability, $S \sim \mathcal{U}(\{ X \mid X \subseteq \{\text{Nu} ,\text{Ac}, \text{M}\} \})$, and only these channels of the images in the batch were passed to the model. Consequently, our model was trained with \textit{multiple views} of the images to be robust to different combinations of ID channels. After training, we found that our model was capable of reconstructing cell-centred tiles from wells not seen during training, as exemplified in Fig. \ref{fig: visualisation}.

\subsection{Self-supervised objective function and channel subsampling leads to biologically informed feature extraction} \label{sec: model vs baseline}
To assess whether \coolcampfire{} distilled useful knowledge from the JUMP-CP dataset, we evaluated our model on several biologically meaningful tasks. In order of increasing complexity, the tasks were to predict the 1-of-9 control compounds of stimulation, using single cell embeddings derived from \begin{inparaenum}[\itshape i\upshape)]
\item plates included in training, or
 \item plates held-out of training, or to predict the 1-of-60 held-out compounds of stimulation using single cell embeddings derived from 
\item  plates included in training, or
\item plates held-out of training. 
\end{inparaenum}

After training \coolcampfire{} we sampled, from all 25 TARGET2 plates, 100 single cell embeddings from each well stimulated by a control compound, and 30 from each well stimulated by a held-out compound. This kept our sample balanced in the representation of wells (due to the low number of cells in some held-out wells).

For the prediction of control compounds, we collected all embeddings derived from training wells stimulated by control compounds, and divided these into 10 equal subsets. For each subset of embeddings, we trained a linear classifier to predict the 1-of-9 control compounds, using a cross-entropy loss function.  Each classifier was trained for 100 epochs, using embeddings from validation wells to perform early stopping with respect to the accuracy. After training, we computed the accuracy of the classifier using embeddings from test wells for the ID and OOD plates, separately. 

For prediction of the held-out compounds, we assigned embeddings derived from held-out compound wells in the ID plates to 5 train/test splits via 5-fold cross-validation. Each fold was used to train a linear classifier for 100 epochs, using early stopping with respect to the accuracy. For each fold, we computed the accuracy with test embeddings from the ID and OOD plates, separately. 

We summarise the performance of our model in Tab. \ref{tab: model vs baselines}. For comparison, we extracted single cell embeddings from the 25 TARGET2 plates using two ImageNet1k baselines and repeated the steps above. We found that across all four tasks, our model outperformed both baseline models, suggesting that our self-supervised objective led to the distillation of useful biological knowledge that can transfer to new tasks.

\begin{table}[b]
\centering
\begin{small}
  \caption{Comparison of three models on four different tasks: predicting compound of stimulation, either ID or OOD, from single cell image embeddings from either ID or OOD plates.  \textbf{DinoViT-S8} and \textbf{DinoViT-L14} are pretrained on ImageNet1k only, \coolcampfire{} was pretrained on JUMP-CP. }
  \label{tab: model vs baselines}
\begin{tabular}{lccccl}\toprule 
      \hline \textbf{Model}  & \thead{\small{ID compound} \\ \small{+ ID plate}} & \thead{\small{ID compound} \\ \small{+ OOD plate}} & \thead{\small{OOD compound} \\ \small{+ ID plate}} & \thead{\small{OOD compound} \\ \small{+ OOD plate}}   \\ 
      \hline \textbf{DinoViT-S8} \citep{caron2021emerging}  & 0.418 ± 0.006  & 0.322 ± 0.007  &0.195 ± 0.006  &0.187  ± 0.003   \\ \hline
       \textbf{DinoViT-L14} \citep{oquab2023dinov2}  &0.356 ± 0.012  &0.281 ± 0.009  &0.158 ± 0.009& 0.152 ± 0.002 \\ 
      \hline \coolcampfire{}  & \textbf{0.460 ± 0.016} & \textbf{0.375 ± 0.007} & \textbf{0.229 ± 0.008}  & \textbf{0.220 ± 0.002}  \\ 
 \bottomrule
\end{tabular}
\end{small}
\end{table}

\subsection{Integration of information from multiple fluorescent channels} \label{sec: channel integration}

\begin{figure*}[t]
\vskip 0.2in
\begin{center}
\includegraphics[width=0.45\linewidth]{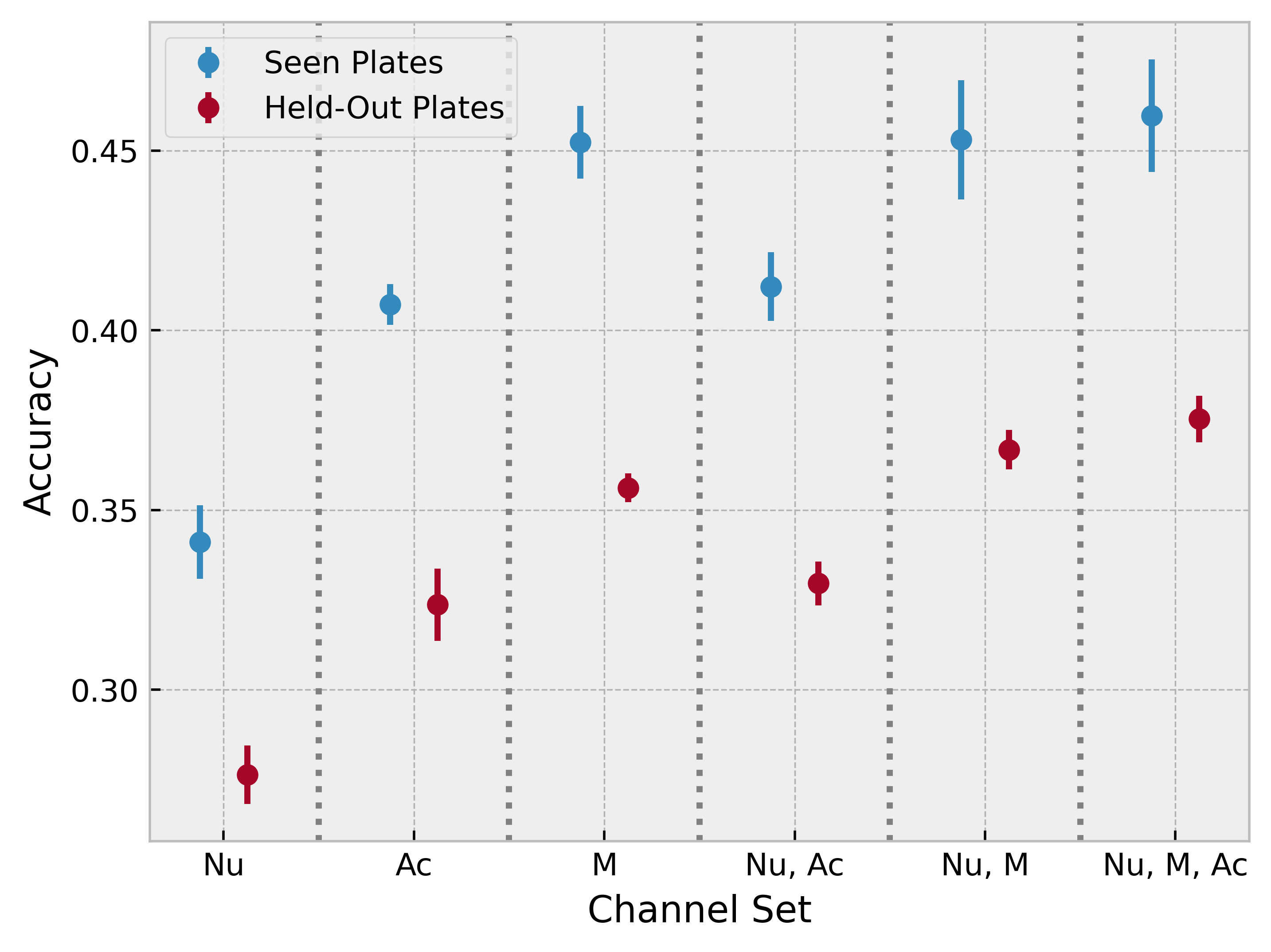}
\includegraphics[width=0.45\linewidth]{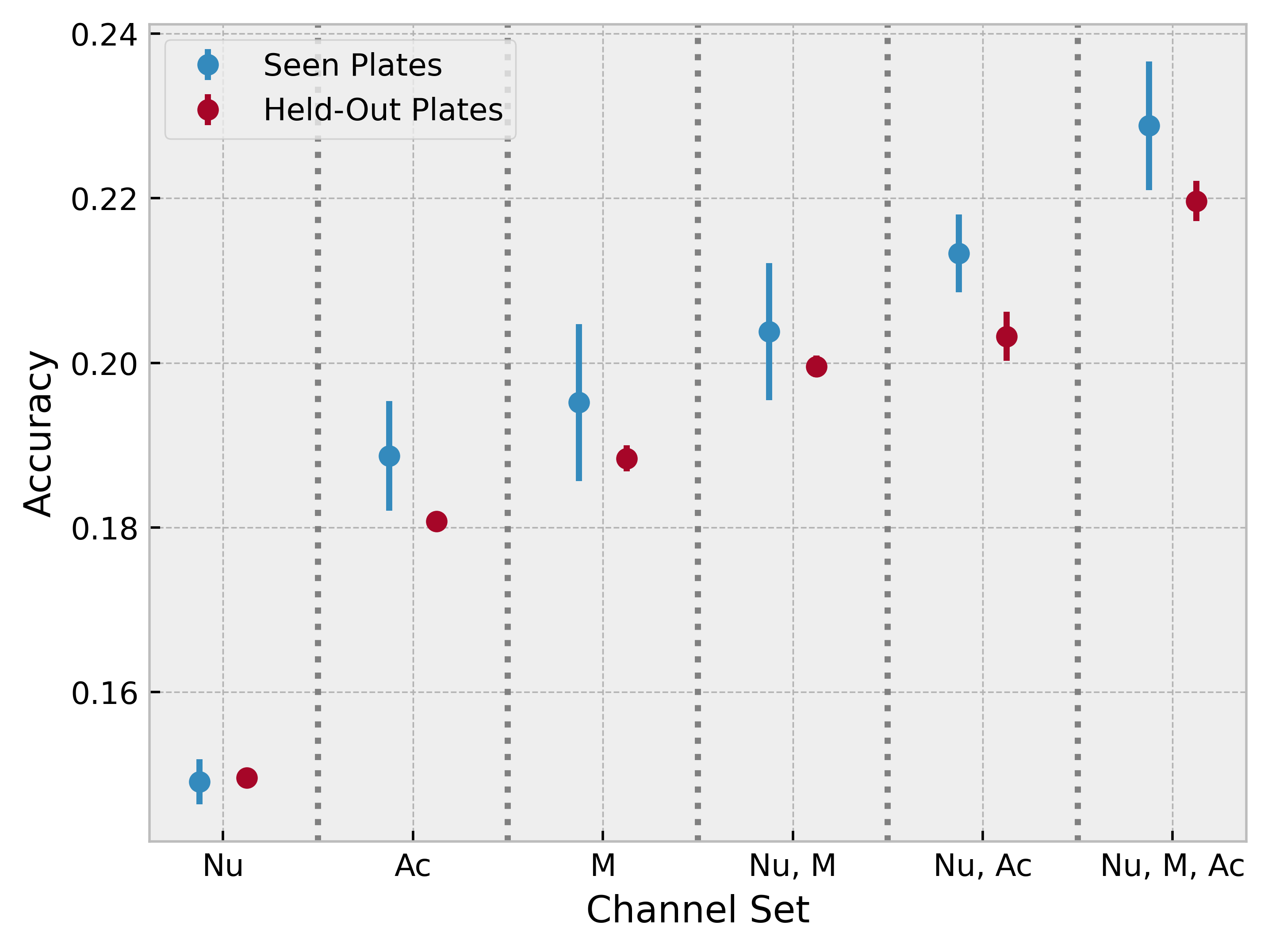}
  \caption{Accuracy of a linear classifier predicting 1-of-9 control compounds (left panel) or 1-of-60 held-out compounds (right) from single cell embeddings. Each column represents a linear classifier trained on single cell embeddings derived from images comprised of different sets of fluorescent channels. Channels shown: Nucleus (N), Actin+Golgi apparatus+Plasma Membrane (Ac), Mitochondria (M) and their combinations.}
  \label{fig: channel integration}
\end{center}
\vskip -0.2in
\end{figure*}

An ideal model for HCI should integrate information from multiple fluorescent channels, such that model embeddings improve with the number of fluorescent channels used to form them. To assess channel integration with \coolcampfire{}, we first extracted single cell embeddings from the 25 TARGET2 plates multiple times, using the following combinations of fluorescent channels,
 \begin{inparaenum}[\itshape i\upshape)]
\item Nu channel only
\item M channel only
\item Ac channel only
\item Nu and Ac channels 
\item Nu and M channels, and 
\item Nu, Ac and M channels.
\end{inparaenum}
For each combination of channels, we extracted single cell embeddings for all cell-centred tiles, hence, extracting embeddings corresponding to \textit{identical} sets of images for each channel combination. We then followed the steps outlined in Sec. \ref{sec: model vs baseline} to sample single cell embeddings and train linear classifiers to predict the compound of stimulation.

In the left panel of Fig. \ref{fig: channel integration}, we show the accuracy of classifiers that predict control compounds. From the accuracy of classifiers trained with a single channel, we found significant difference in the quality of information that each channel provided. The M channel was found to be the most informative channel, alone. It was always found beneficial to include additional fluorescent channels; the Nu channel and Ac channel in combination provided higher accuracy than the Ac channel alone, and despite the Nu channel being the least informative channel, the Nu and M channels in combination provided higher accuracy than the M channel alone. 

We compared this with the accuracy of classifiers trained to predict the held-out compounds in the right panel of Fig. \ref{fig: channel integration}. Comparing the rank of each combination of channels, in terms of their predictive accuracy, we found that the rank was not consistent between the tasks of predicting control and held-out compounds. Although the rank of single channel accuracy was consistent, we found that the combination of the Nu and Ac channels outperformed the combination of the Nu and M channels, in the prediction of the held-out compounds, despite M being the single most informative channel. This suggests that the interaction between fluorescent channels impacted the quality of image embeddings, and that embedding quality is not maximised by merely selecting the single most informative channels. From this we concluded that our model successfully integrates information across fluorescent channels.

\subsection{Generalisation to held-out fluorescent channels}

To evaluate generalisation to OOD fluorescent channels, we extracted single cell embeddings from the 25 TARGET2 plates using the Nu, ER and cyRNA channels (although the Nu channel was ID, we included it in the OOD channel set as it is unlikely for a nucleus channel to be excluded in a screen). We sampled single cell embeddings and trained linear classifiers according to the steps in Sec. \ref{sec: model vs baseline}.


We compared the performance of \coolcampfire{} to our strongest baseline \textbf{DinoViT-S8} (for which all channels are OOD, as it is pretrained on ImageNet1k only). In Tab. \ref{tab: unseen channel}, the accuracy of \textbf{DinoViT-S8} was found to increase with OOD channels across all 4 tasks. Having seen that single channels provide different amounts of information in Sec. \ref{sec: channel integration}, this implied that an informative channel was held-out during training. Conversely, when comparing ID to OOD channels, \coolcampfire{} showed lower accuracy on all 4 tasks, as is expected due to distribution-shift. Despite this, \coolcampfire{} outperformed  \textbf{DinoViT-S8} on the task of highest difficulty, predicting the held-out compound from OOD plates, suggesting that \coolcampfire{} distilled useful biological knowledge that transfers to OOD fluorescent channels. 

\begin{table}[t]
\centering
\begin{small}
  \caption{\small{Comparison of 2 models on 4 different biological tasks, predicting compound of stimulation, either ID or OOD, from single cell image embeddings from either ID or OOD plates. Results are shown when inference is performed with either ID or OOD sets of channels. \textbf{DinoViT-S8} was pretrained on ImageNet1k, \coolcampfire{} was pretrained on JUMP-CP data, with the ID channel set. Metric is the accuracy shown over 10 training splits (ID compound) or 5 folds (OOD compound), with ± indicating standard deviation.  }}
  \label{tab: unseen channel}
\begin{tabular}{l|ccccl}\toprule 
      \hline \textbf{Model / Channel Set}  & \thead{\small{ID compound} \\ \small{+ ID plate}} & \thead{\small{ID compound} \\ \small{+ OOD plate}} & \thead{\small{OOD compound} \\ \small{+ ID plate}} & \thead{\small{OOD compound} \\ \small{+ OOD plate}}  \\ \hline
      \textbf{Campfire / ID}  & \textbf{0.460 ± 0.016} & \textbf{0.375 ± 0.007} & \textbf{0.229 ± 0.008}  & \textbf{0.220 ± 0.002}  \\
      \textbf{DinoViT-S8 / ID}  & 0.418 ± 0.006  & 0.322 ± 0.007 & 0.195 ± 0.006  & 0.187 ± 0.003  \\
      \midrule
       \textbf{Campfire / OOD}  & 0.44 ± 0.01 & 0.353 ± 0.07 & \textbf{0.206 ± 0.009} & \textbf{0.199 ± 0.003}   \\ 
        \textbf{DinoViT-S8 / OOD}  & \textbf{0.457 ± 0.009}  & \textbf{0.356 ± 0.009} &  0.204 ± 0.003 &  0.191 ± 0.001   \\ 
 \bottomrule
\end{tabular}
\end{small}
\end{table}

\subsection{Transfer learning from one microscopy screen to another}\label{sec: finetuning}

To assess the extent that \coolcampfire{} could transfer knowledge from the JUMP-CP dataset to a different microscopy screen, we finetuned \coolcampfire{} to detect changes in macrophage morphology. We considered a dataset comprised of 4, 384-well, plates containing macrophages subject to different gene knock-out \citep{mehrizi2023multi}. Two plates contained macrophages with M1 polarisation, while the other 2 contained macrophages with M2 polarisation. We froze the parameters of \coolcampfire{} and attached a 2-layer MLP (with hidden dimensions 1024 and 128) to its last layer. The resultant model was then trained using a triplet loss objective function \citep{schroff2015facenet} using one M1 plate and one M2 plate, leaving the other two plates held-out of training for later evaluation. For each mini-batch we sampled several cell-centred tiles from the same set of wells, and computed the mean model embedding for each well in the batch. These \textit{well-level} embeddings were then passed to the triplet loss objective function, where we treated well-level embeddings derived from wells subject to the same/different gene knock-out as positive/negative samples. After 500 epochs of training, we ran inference over all four plates, and computed the well-level embedding for each well. We repeated this process for the \textbf{DinoViT-S8} model.

Our dataset contained wells subject to positive and negative control stimulations that had, respectively, a known impact or lack of impact on macrophage morphology. We computed the $Z'$-score \citep{zhang1999simple} for each pairwise combination of positive and negative controls (calculation detailed in App. \ref{app: zprime}) . This score indicated the size of the statistical difference between the groups of embeddings subject to two different stimulations. 

\begin{figure*}[t]
\vskip 0.2in
\begin{center}
\includegraphics[width=0.45\linewidth]{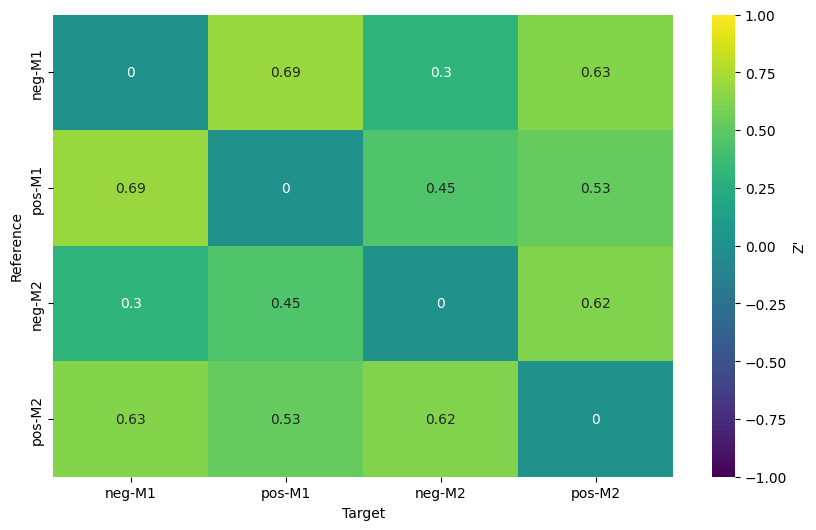}
\includegraphics[width=0.45\linewidth]{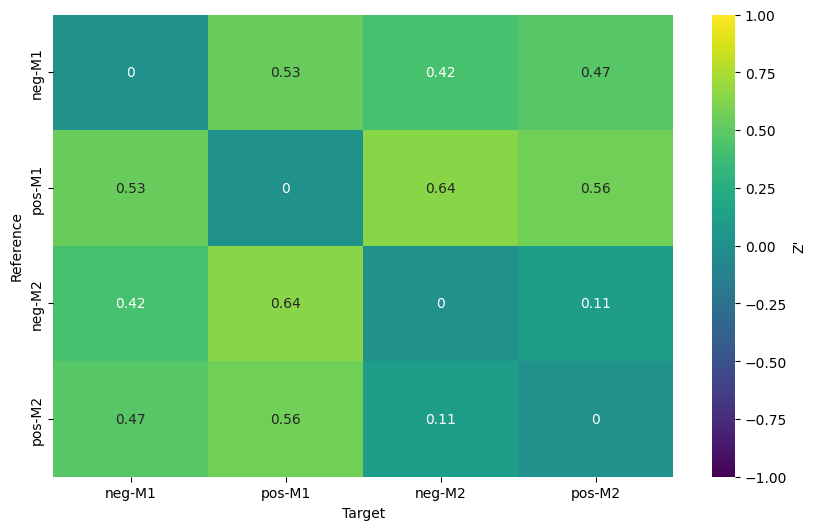}
\caption{$Z'$-score measuring statistical difference between model embeddings from a reference and target compound of stimulation. Model embeddings are derived from \coolcampfire{} (left) pretrained on JUMP-CP, and \textbf{DinoViT-S8} (right) pretrained on ImageNet1k. Both models have been finetuned on a macrophage dataset. $Z'$-score is shown for negative (neg) and positive (pos) controls for plates with macrophages in either M1 or M2 polarisation.  }

  \label{fig: macrophage zprime}
\end{center}
\vskip -0.2in
\end{figure*}

As shown in Fig. \ref{fig: macrophage zprime}, we found that \coolcampfire{} was demonstrably better than \textbf{DinoViT-S8} at distinguishing between M1 macrophages that had been subject to positive or negative controls. To a greater extent, this was also true for M2 macrophages, for which \textbf{DinoViT-S8} failed to discern between stimulations. Despite this difference, the models were comparable when discerning between macrophages of different polarisation, regardless of stimulation. That \textbf{DinoViT-S8} could discern between macrophages of different polarisation, which were derived from different plates, but not between positive and negative controls, suggested it was prone to plate effects. From this we concluded that pretraining on HCI allowed \coolcampfire{} to mitigate batch effects without additional training constraints during finetuning. 

\FloatBarrier

\section{Conclusion}

Developing a channel-agnostic model that can generalise to new experimental plates, compounds of perturbation, OOD fluorescent channels and cell types is essential if a foundation model for HCI is to be realised. The development of a foundation model for HCI not only requires specialised model design and dataset curation, but also proper evaluation, to ascertain how choices in architecture, dataset, and training protocol affect the generalisation of models for HCI. In this work we proposed a method of splitting the JUMP-CP dataset that allowed us to scrutinise how model design and training protocol impacted the ability of a model to generalise under a specific source of distribution-shift. 

We developed a channel-agnostic model for HCI that, when trained in self-supervised fashion, could predict the compound of stimulation from images at single cell resolution. Our model, \coolcampfire{}, generalised to OOD experimental batches, compounds, and sets of fluorescent channels. Of these sources of distribution-shift, generalising to a new set of fluorescent channels was found to be the most challenging. All images in the JUMP-CP dataset consisted of 5 fluorescent channels which limited the number of channels that could be included in training, in order to form an appropriate OOD evaluation set. Training over a broader set of fluorescent channels would likely have improved the generalisation of our model under this source of distribution-shift. While we trained \coolcampfire{} with the JUMP-CP dataset exclusively, training with screens of different cell types will be crucial for a HCI foundation model. This may also improve generalisation to OOD fluorescent channels, as ID fluorescent markers will capture broader variation due to differences in the morphology of different cell types. Despite the challenges of generalising to OOD channels, \coolcampfire{} successfully demonstrated transfer learning to a microscopy screen containing new cell types. Our evaluation scheme allows us to compare model performance with respect to different distribution shifts, and compare this with their performance when transfer learning to a new microscopy screen, where these models would be used in practice, hopefully increasing the rate at which a foundation model for HCI can be realised. 

While our work focussed on predicting the compound of perturbation, our proposed method to split the wells of the JUMP-CP dataset into training, validation and test sets may be used to evaluate model performance on different biologically relevant tasks, while still isolating the different sources of distribution-shift we considered. Future work towards the development of a universal model for HCI should focus on leveraging data from multiple sources and increasing the number of fluorescent channels and cell types included in training, in order to produce high quality embeddings under realistic sources of variation in images produced by HCS. 

\subsubsection*{Meaningfulness Statement}

Life as we know it is a multi-scale phenomena, from the structure of DNA, to cell state, tissue function, and organ interactions.
A meaningful representation of life should capture how changes at one scale affect another. Our work focusses on how perturbations at the sub-cellular level affect cell phenotype. By evaluating channel agnostic models for high content screening with respect to different distribution shifts, we hope to aid the development of a foundation model for high content imaging which provides robust representations of cells subject to perturbation that capture fine-grained information of cell phenotype.

\bibliography{campfire}
\bibliographystyle{iclr2025_conference}
\appendix

\section{Model objective function} \label{app: loss}

During pretraining \coolcampfire{} was trained to optimise the following objective function, 
\begin{eqnarray}
\mathcal{L} = - \frac{1}{|\mathcal{D}|}\sum_{i=1}^{|\mathcal{D}|} \left[ \lambda_{s} MSE(\hat{y}_i,y_{i}) + \lambda_{\ell} MSE(\mathcal{F}[\hat{y}_i| \ell],\mathcal{F}[y_{i} |\ell ]) + \lambda_{h} MSE(\mathcal{F}[\hat{y}_i| h],\mathcal{F}[y_{i} |h])\right], \label{eq: mae loss}
\end{eqnarray}
where $y_i$ and $\hat{y}_i$ are the unmasked image, and image reconstruction, respectively, and $MSE(x_1,x_2) = \frac{1}{M}\sum_{j=1}^{M} (x_{1,j} - x_{2,j})^{2}$ for $x_{1},x_2 \in \mathcal{R}^{M}$ is the L2 reconstruction loss and $|\mathcal{D}|$ is the number of samples in the dataset.  The hyperparameters $\lambda_{s,l,h}\in[0,1]$ control the weight given to each term in the loss function. $F[x|h/\ell ]$ represents a high/low pass filter function. The high pass filter was performed by applying a 2D fast Fourier transform (FFT), setting the outer-most Fourier coefficients to zero, and applying the inverse 2D FFT. Similarly, the low-pass filter involved applying the 2D FFT, removing the inner-most Fourier coefficients and performing the inverse 2D FFT. Here  $h,\ell \in [0,1]$ represent the fraction of Fourier coefficients removed in the high and low pass filters, respectively. Consequently, our model was trained to reconstruct uncorrupted images, and also optimise the reconstruction of images subject to high and low pass filters, as in recent works \citep{juyal2024pluto,kraus2024masked}.

\section{Ablation study} \label{app: ablation}

In order to determine the optimum hyperparameters of our model we performed a three-stage ablation study which sought to optimise hyperparameters for: \begin{inparaenum}[\itshape i\upshape)]
\item the objective function,
\item neural network architecture, and
\item optimiser.
\end{inparaenum}

In the first stage of the ablation study, we optimised the hyperparameters of the loss function
\begin{eqnarray}
\mathcal{L}_{i} &=& - \mathcal{L}_{spatial,i}- \mathcal{L}_{filter,i} - \mathcal{L}_{frequency,i}  \\
\mathcal{L}_{spatial,i} &=&  \lambda_{s} MSE(\hat{y}_i,y_{i})  \\
\mathcal{L}_{filter,i} &=&  \lambda_{h} MSE(\mathcal{F}[\hat{y}_i| h],\mathcal{F}[y_{i} |h])  + \lambda_{\ell} MSE(\mathcal{F}[\hat{y}_i| \ell],\mathcal{F}[y_{i} | \ell])   \\
\mathcal{L}_{frequency,i} &=&   \lambda_{f} L_{1}(FFT(\hat{y}_i), FFT(y_i) ).
\end{eqnarray}
Here $y_i$ and $\hat{y}_i$ are the input image and the reconstructed image, respectively. The 2D fast Fourier transform (FFT) of an image is represented here by $FFT(y_i)$. The per-sample loss function $\mathcal{L}_i$ was a weighted combination of $\mathcal{L}_{spatial,i}$ which optimised the model for reconstruction of the input images, $\mathcal{L}_{filter,i} $ which optimised the reconstruction of the images filtered for high and low frequencies, and $\mathcal{L}_{frequency,i}$ which optimised the model for reconstruction of images in the frequency domain. The weights of each loss term were controlled by $\lambda_s$, $\lambda_f$ and $\lambda_h$ and $\lambda_l$.

 We performed hyperparameter optimisation via grid search, training a model for different configurations of the hyperparameters in the above loss function, with the values stated in Tab. \ref{tab: loss ablation}. For each trial, we trained a model for 50 epochs on 2 A100 GPUs with a global batch size of 512, AdamW optimiser with L2 regularisation $0.002$, a linearly increasing learning rate in the first 10 epochs from $1\times 10^{-5}$ to $1 \times 10 ^{-4}$, followed by cosine annealing for the next 40 epochs from $1\times 10^{-4}$ to $1 \times 10 ^{-6}$. The patch size of the model was set to $(16,16)$, cell-centred tiles were sized at (112,112), patches were masked during training with fraction $p_m = 0.75$, and we trained the model using the 20 TARGET2 plates only, excluding the COMPOUND plates from the hyperparameter optimisation. We selected the best configuration based on the validation loss and validation reconstruction error, and highlight this configuration in Tab. \ref{tab: loss ablation}.

 After finding the optimal configuration for the hyperparameters for the loss function, we performed the second stage our hyperparameter optimisation, focussing on hyperparameters concerning the neural network architecture. In this stage of hyperparameter optimisation, we varied the size of the patch size of the cell-centred tiles, whether or not we initialised our encoder with ImageNet1k weights, the size of our encoder (from the standard sizes Large or Huge provided by HuggingFace), the rate of stochastic depth in the attention blocks (Drop Path Rate), and whether patches were masked at random (Sync Mask is False) or patches are masked such that patches from the same position but different channel are masked together (Sync Mask is True). We trained a model for each of the configurations of these hyperparameters in Tab. \ref{tab: model ablation} for 50 epochs using 2 A100 GPUs with a global batch size of 128,  and AdamW optimiser with L2 regularisation $0.002$, a linearly increasing learning rate in the first 10 epochs from $1\times 10^{-5}$ to $1 \times 10 ^{-4}$, followed by cosine annealing for the next 40 epochs from $1\times 10^{-4}$ to $1 \times 10 ^{-6}$. We used the optimal loss hyperparameters from the first stage of hyperparameter optimisation. We highlight the optimal configuration in Tab. \ref{tab: model ablation}. We then perturbed the $p_m$ alone, and found that increasing the mask ratio $p_m=0.8$ led to higher performance. 
 
 In the last stage of our ablation analysis, we optimised the learning rate and weight decay of our AdamW optimiser. We trained a model for each of the configurations of these hyperparameters in Tab. \ref{tab: optimiser ablation} for 50 epochs using 2 A100 GPUs with a global batch size of 400,  and AdamW optimiser with L2 regularisation $0.002$, a linearly increasing learning rate in the first 10 epochs from $1\times 10^{-5}$ to $1 \times 10 ^{-4}$, followed by cosine annealing for the next 40 epochs from the learning rate start value for the given trial to $1 \times 10 ^{-6}$. The optimal configuration is shown in Tab. \ref{tab: optimiser ablation}.

\begin{table}[h]
\centering
\begin{tiny}
  \caption{\small{ Hyperparameter combinations tested during ablation study for loss function. Each row corresponds to a model trained with the stated configuration of hyperparameters. Best configuration is highlighted in bold. }}
  \label{tab: loss ablation}
\begin{tabular}{lccccccl}\toprule 
    Trial & $\lambda_{s}$ & $\lambda_{h }$ & $\lambda_{l}$ & $\lambda_{f}$  & $h$ & $\ell$  \\ \hline
         1  & 1.0  & 0.00 & 0.00 & 0.00 & NA   & NA   \\
         2  & 0.75 & 0.00 & 0.00 & 0.25 & NA   & 0.1  \\
         3  & 0.75 & 0.00 & 0.25 & 0.00 & 0.1  & NA   \\
         4  & 0.34 & 0.00 & 0.33 & 0.33 & 0.1  & 0.1  \\
         5  & 0.25 & 0.00 & 0.50 & 0.25 & 0.1  & 0.1  \\
         6  & 0.25 & 0.00 & 0.25 & 0.50 & 0.1  & 0.1  \\
         7  & 0.99 & 0.01 & 0.00 & 0.00 & NA   & NA   \\
         8  & 0.74 & 0.01 & 0.00 & 0.25 & NA   & 0.1  \\
         9  & 0.74 & 0.01 & 0.25 & 0.00 & 0.1  & NA   \\
        10  & 0.33 & 0.01 & 0.33 & 0.33 & 0.1  & 0.1  \\
        11  & 0.24 & 0.01 & 0.50 & 0.25 & 0.1  & 0.1  \\
        12  & 0.24 & 0.01 & 0.25 & 0.50 & 0.1  & 0.1  \\
        13  & 0.50 & 0.00 & 0.50 & 0.00 & 0.1  & NA   \\
        14  & 0.25 & 0.00 & 0.75 & 0.00 & 0.1  & NA   \\
        15  & 0.00 & 0.00 & 1.00 & 0.00 & 0.1  & NA   \\
        16  & 0.75 & 0.00 & 0.25 & 0.00 & 0.05 & NA   \\
        17  & 0.50 & 0.00 & 0.50 & 0.00 & 0.05 & NA   \\
        18  & 0.25 & 0.00 & 0.75 & 0.00 & 0.05 & NA   \\
        19  & 0.00 & 0.00 & 1.00 & 0.00 & 0.05 & NA   \\
        20  & 0.90 & 0.00 & 0.10 & 0.00 & 0.1  & NA   \\
        21  & 0.90 & 0.00 & 0.10 & 0.00 & 0.3  & NA   \\
        22  & 0.90 & 0.00 & 0.10 & 0.00 & 0.5  & NA   \\
        23  & 0.90 & 0.00 & 0.10 & 0.00 & 0.7  & NA   \\
        24  & 0.75 & 0.00 & 0.25 & 0.00 & 0.2  & NA   \\
        25  & 0.75 & 0.00 & 0.25 & 0.00 & 0.25 & NA   \\
        \textbf{26} & \textbf{0.75} & \textbf{0.00} & \textbf{0.25} & \textbf{0.00} & \textbf{0.3}  & \textbf{NA}   \\
        27  & 0.75 & 0.00 & 0.25 & 0.00 & 0.35 & NA   \\
 \bottomrule
\end{tabular}
\end{tiny}
\end{table}

\begin{table}[h!]
\centering
\begin{tiny}
  \caption{\small{ Hyperparameter combinations tested during ablation study for neural network architecture. Each row corresponds to a model trained with the stated configuration of hyperparameters. Best configuration is highlighted in bold.  }}
  \label{tab: model ablation}
\begin{tabular}{lccccccl}\toprule 
    Trial & Patch Size & $p_{m }$ & Drop Path Rate  &  Sync Mask   & ImageNet1k weights & Encoder Size  \\ \hline
          1 & 16 & 0.75 & 0 & True & True & Large \\
        \textbf{2} & \textbf{14} & \textbf{0.75} & \textbf{0} & \textbf{True} & \textbf{True} & \textbf{Large} \\
        3 & 8 & 0.75 & 0 & True & True & Large \\
        4 & 16 & 0.75 & 0.1 & True & True & Large \\
        5 & 14 & 0.75 & 0.1 & True & True & Large \\
        6 & 8 & 0.75 & 0.1 & True & True & Large \\
        7 & 16 & 0.75 & 0.2 & True & True & Large \\
        8 & 14 & 0.75 & 0.2 & True & True & Large \\
        9 & 8 & 0.75 & 0.2 & True & True & Large \\
        10 & 16 & 0.75 & 0.3 & True & True & Large \\
        11 & 14 & 0.75 & 0.3 & True & True & Large \\
        12 & 8 & 0.75 & 0.3 & True & True & Large \\
        13 & 16 & 0.75 & 0 & True & False & Large \\
        14 & 14 & 0.75 & 0 & True & False & Large \\
        15 & 8 & 0.75 & 0 & True & False & Large \\
        16 & 16 & 0.75 & 0 & False & True & Large \\
        17 & 14 & 0.75 & 0 & False & True & Large \\
        18 & 8 & 0.75 & 0 & False & True & Large \\
        19 & 14 & 0.75 & 0 & True & True & Huge \\
        20 & 16 & 0.75 & 0 & True & True & Huge \\
        21 & 14 & 0.75 & 0 & True & False & Huge \\
        22 & 16 & 0.75 & 0 & True & False & Huge \\
 \bottomrule
\end{tabular}
\end{tiny}
\end{table}

\begin{table}[h!]
\centering
\begin{tiny}
  \caption{\small{ Hyperparameter combinations tested during ablation study for the optimiser. Each row corresponds to a model trained with the stated configuration of hyperparameters. Best configuration is highlighted in bold.  }}
  \label{tab: optimiser ablation}
\begin{tabular}{lcccl}\toprule 
    Trial  & Learning Rate & Weight Decay \\ \hline
        1 & 1.0e-4 & 2.0e-3 \\
        2 & 5.0e-5 & 2.0e-3 \\
        3 & 1.5e-4 & 2.0e-3 \\
        4 & 2.0e-4 & 2.0e-3 \\
        5 & 1.0e-4 & 5.0e-3 \\
        6 & 1.0e-4 & 1.0e-2 \\
        7 & 5.0e-4 & 2.0e-3 \\
        8 & 1.0e-4 & 1.0e-3 \\
        9 & 3.0e-4 & 5.0e-3 \\
        10 & 4.0e-4 & 5.0e-3 \\
        \textbf{11} & \textbf{5.0e-4} & \textbf{5.0e-3} \\
        12 & 6.0e-4 & 5.0e-3 \\
 \bottomrule
\end{tabular}
\end{tiny}
\end{table}

\FloatBarrier

\section{$Z'$-score from model embeddings} \label{app: zprime}

The $Z'$-score is typically computed for scalar values. Here we describe the steps taken to compute the $Z'$-score to measure the statistical size effect between different groups of model embeddings, corresponding to different compound stimulations. 

The dataset we finetuned \coolcampfire{} with in Sec. \ref{sec: finetuning} was comprised of plates with wells that were treated with either negative or positive controls. 
Control compounds, with a priori negative or positive effect on cell morphology, can be used to measure assay quality, but were here used to evaluate model embeddings. From the M1 and M2 plates we computed the well-level embeddings for each well stimulated by positive or negative controls. These wells were labelled $\{\text{neg-M1, neg-M2, pos-M1, pos-M2}\}$. 

Our aim was to compute the $Z'$ score for each pairwise combination of these labels. To do so we converted our model embeddings to a single value corresponding to the difference in the clusters of embeddings for each label. Given two groups of embeddings,  $X$ and $Y$ with labels $\ell_x$ and $\ell_y$, we treated $X$ as the \textit{reference} group, and $Y$ as the \textit{target} group. We computed the mean embedding from the group $X$, $\mu_x$, and subtracted this from all embeddings in both groups i.e $X = X-\mu_x$ and $Y = Y-\mu_x$. We then computed the mean embedding of $Y$, denoted $\mu_y$, and for each embedding in $X$ and $Y$ we computed the normalised scalar projection on to $\mu_y$. Hence, for the target and reference group, each embedding was represented by a single value quantifying its alignment with the target group. We computed the mean, $\mu_t$ and $\mu_r$, and standard deviation, $\sigma_t$ and $\sigma_r$, of these values for the target ($t$) and reference ($r$) group and then computed $Z'$ using the formula, 
\begin{equation}
Z' = 1 - 3 \frac{\sigma_r + \sigma_t }{|\mu_r - \mu_t|}.
\end{equation}

\end{document}